\documentclass[10pt,twocolumn,letterpaper]{article}

\usepackage[pagenumbers]{cvpr} 
\usepackage{amsthm}
\usepackage[ruled, linesnumbered, vlined]{algorithm2e}
\theoremstyle{plain}
\newtheorem{theorem}{Theorem}[section]

\newtheorem{lemma}[theorem]{Lemma}

\newtheorem{assumption}{Assumption}
\definecolor{cvprblue}{rgb}{0.21,0.49,0.74}
\usepackage[pagebackref,breaklinks,colorlinks,allcolors=cvprblue]{hyperref}

\def\confName{CVPR}
\def\confYear{2027}

\title{Dune: A Zeroth-Order Optimizer with Time-Conditioned Historical Error Regression for VLA Test-Time Adaptation}

\author{%
  Chengyang Dong\textsuperscript{1,4,7,*},
  Ze Chen\textsuperscript{2,*},
  Kai Ye\textsuperscript{5},
  Chutian Wang\textsuperscript{3},
  Heng Xu\textsuperscript{4}
  Wei Mo\textsuperscript{6},\\
  Maomao Li\textsuperscript{7},
  Jiayan Xu\textsuperscript{8},
  Jiayuan Wu\textsuperscript{9},
  Yihao Xiao\textsuperscript{10}
  Linhua Ye\textsuperscript{11},\\
  Zhonghua Peng\textsuperscript{12},
  Ruize Han\textsuperscript{4},
  Jialong Sun\textsuperscript{13,14,\ensuremath{\dagger}}\\[4pt]
  \textsuperscript{1}Statistics and Mathematics, University of Glasgow
  \textsuperscript{2}Beijing Normal University\\
  \quad \textsuperscript{3}Sun Yat-sen University
  \textsuperscript{4}Shenzhen University of Advanced Technology\\
  \textsuperscript{5}School of Artificial Intelligence, Shanghai Jiao Tong University\\
  \textsuperscript{6}Faculty of Electrical and Computer Engineering,
  Technische Universit\"at Dresden\\
  \textsuperscript{7}The University of Hong Kong
  \textsuperscript{8}Southern University of Science and Technology\\
  \textsuperscript{9}Tongji University
  \textsuperscript{10}Shanghai University of Finance and Economics\\
  \textsuperscript{11}South China University of Technology
  \quad \textsuperscript{12}Shenzhen University\\
  \textsuperscript{13}School of Cybersecurity, Heilongjiang University
  \quad \textsuperscript{14}Shandong University\\%
  {\small
    \textsuperscript{*}These authors contributed equally.
    \quad \textsuperscript{\ensuremath{\dagger}}Corresponding author.%
  }
  \\[4pt]{\small\texttt{jialong.sunh@gmail.com}}
}

\begin{document}
\maketitle
\begin{abstract}
Test-Time Adaptation (TTA) enables Vision-Language-Action (VLA) models to handle distribution shifts during deployment without source-domain retraining. However, backpropagation-based adaptation incurs heavy memory overhead on resource-constrained robots, whereas zeroth-order (ZO) methods suffer from high estimation variance under limited perturbation budgets. We observe that while ZO estimation enables low-memory adaptation with rapid initial gains, it often stalls or degrades over long-horizon updates due to noisy gradient estimates. To address these limitations, we propose Dune, a zeroth-order optimizer that stabilizes long-horizon adaptation by coupling online historical error prediction with current perturbation observations. Specifically, Dune learns a lightweight, time-conditioned error predictor from past optimization dynamics and fuses its predictions with instantaneous ZO estimates, achieving effective variance reduction without backpropagation or extra forward passes. Additionally, we theoretically characterize how estimation variance restricts convergence in high-dimensional, small-population regimes, motivating the use of historical statistics. Extensive experiments on LIBERO and real-world robots demonstrate that Dune achieves superior success rates under a low memory footprint, with ablations confirming improved closed-loop stability and post-disturbance recovery.
\end{abstract}    
\section{Introduction}
Recent vision-language-action (VLA) models translate visual observations and language instructions into effective actions across a wide range of robotic manipulation and embodied control tasks~\cite{kim2024openvla,black2024pi_0,brohan2022rt,brohan2023rt,li2026safetyembodiedaisurvey,chen2026hazardarenaevaluatingsemanticsafety}. In dynamic environments, agents must reason over long spatiotemporal horizons and adapt their actions as conditions change. Most VLA models rely on large-scale offline pretraining~\cite{intelligence2025pi,liu2023libero,o2024open,khazatsky2024droid,lin2026evodepthlightweightdepthenhancedvisionlanguageaction}. They perform well on in-distribution tasks but struggle in out-of-distribution (OOD) settings, where they encounter tasks and environmental variations that pretraining data rarely cover~\cite{black2024pi_0,intelligence2025pi,wang2026libero}.

Test-time adaptation (TTA) offers a way for VLA models to respond to distribution shifts during deployment~\cite{sun2020test,bai2025evolve,park2026test,liu2026fly}. As a robot acts, incoming visual observations and the resulting environmental feedback provide supervision without expert action labels~\cite{ji2026vane}. These signals can help the model adapt its perceptual representations and action decisions while retaining knowledge acquired during pretraining. On-device fine-tuning with first-order (FO) optimization remains difficult because the memory and compute demands of backpropagation (BP) can exceed the resources available on robotic platforms, particularly those that use low-bit quantized weights for inference~\cite{dettmers2023qlora,malladi2023fine}.

\begin{figure}[t]
    \centering
    \includegraphics[width=1\linewidth]{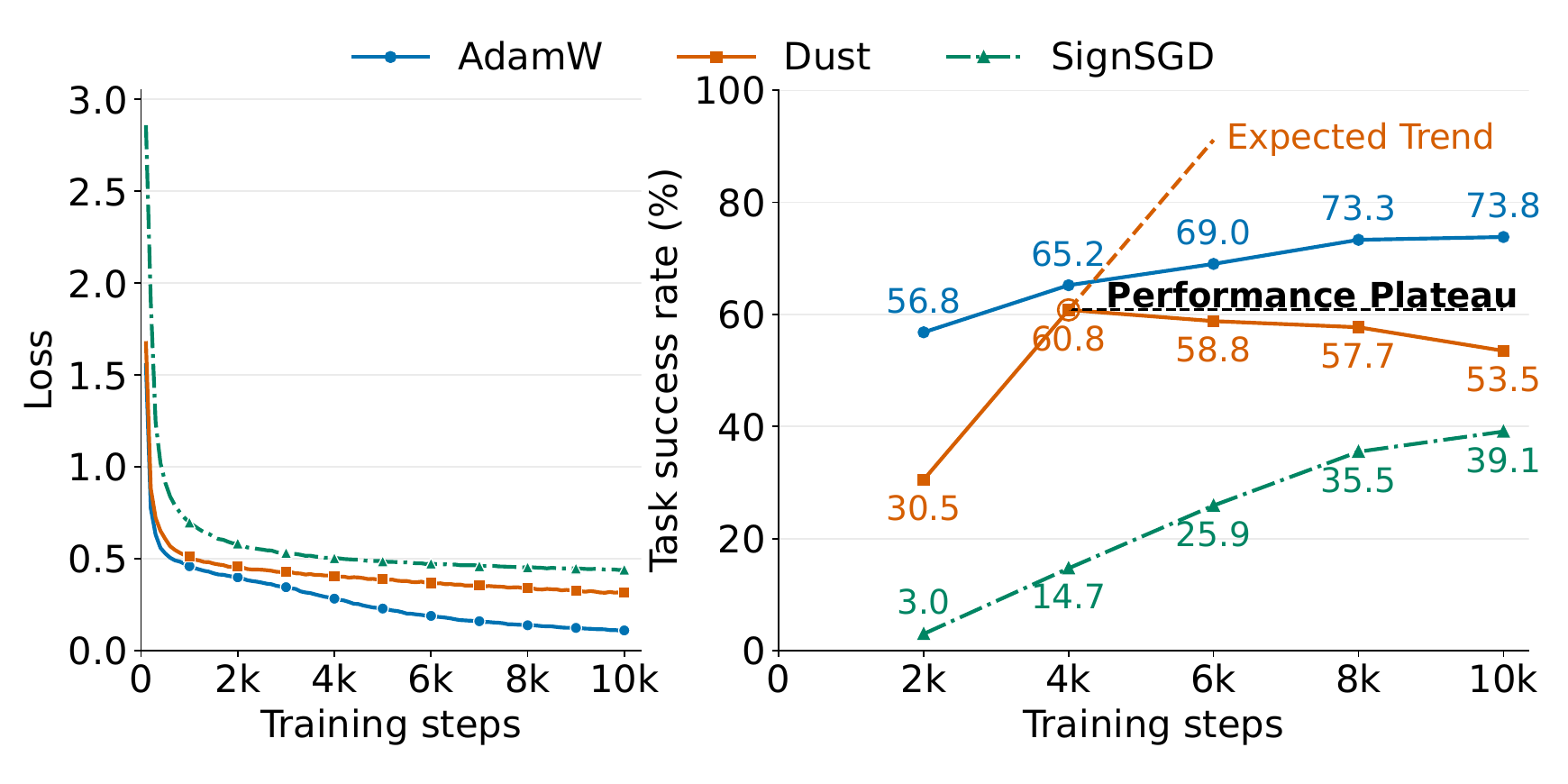}
    \caption{Loss and task success rate in a preliminary comparison of Dust and standard first-order optimizers for VLA test-time adaptation (VLA-TTA). Dust uses a perturbation population size of $K=64$ and converges faster early in adaptation. After 6{,}000 steps, its task success rate plateaus and then declines, while the first-order optimizers continue to improve steadily.}
    \label{fig:placeholder}
\end{figure}

Zeroth-order (ZO) optimization reduces the memory cost of TTA by estimating gradients from forward evaluations under perturbations~\cite{berahas2021theoreticalempiricalcomparisongradient,wang2026fozo,zhang2026curvature,zhang2026zotta}. MeZO uses this approach to fine-tune large language models with near-inference memory usage and achieves performance comparable to first-order fine-tuning on several downstream tasks~\cite{malladi2023fine}. Dust, a state-of-the-art ZO optimizer, uses activation-space perturbations and parallel evaluations to match or exceed the performance of backpropagation at large perturbation population sizes~\cite{dahal2026backprop}. In our preliminary VLA-TTA experiment (Figure~\ref{fig:placeholder}), Dust quickly approaches AdamW's task success rate~\cite{kingma2014adam,loshchilov2017decoupled}. After 6{,}000 steps, Dust's success rate plateaus and then drops substantially, eventually reaching a level comparable to that of SignSGD. We refer to this pattern of stagnation followed by decline during long-horizon ZO adaptation as the \textbf{performance plateau problem}.

The performance plateau problem motivates \textbf{Dune}   (Dust Using Nodewise Error-feedback), a memory-efficient ZO optimizer that uses regression over historical statistics to recover reliable update directions when perturbation noise overwhelms the dominant error signal. At each step, Dune constructs regression features from model output residuals and time buckets that encode the stage of adaptation. It uses historical statistics with decaying weights to predict layerwise error signals from these features. Dune then forms a weighted combination of these predictions and the current ZO perturbation observations to preserve the dominant direction of the error signal and suppress high-frequency perturbation noise. Finally, it updates the historical statistics with the raw perturbation observations from the current batch, allowing the regression to track changes as adaptation proceeds. This mechanism stabilizes long-horizon VLA-TTA while retaining the memory efficiency of ZO optimization. Our paper mainly makes the following contributions:
\begin{itemize}
   
 \item \textbf{Dune Paradigm.} To overcome performance plateaus in VLA-TTA, Dune predicts layer-wise error signals via online ridge regression over output residuals and flow-matching time buckets, fusing them with current perturbation observations for stable updates without backpropagation or extra forward passes.

\textbf{Performance Plateau Theory.} Under Gaussian perturbation and convex quadratic objectives, we establish a loss recursion showing how estimation variance limits convergence in high-dimensional, small-population regimes, theoretically motivating historical error regression.

\textbf{Extensive Experiments.} Across LIBERO benchmarks and real-world robots, Dune substantially boosts long-horizon task success rates with a significantly reduced memory footprint, while ablations confirm improved closed-loop stability and post-disturbance recovery.
\end{itemize}

\section{Related Work}
\textbf{Vision--language--action models.} Large-scale robot datasets and multimodal pretraining support the learning of general-purpose manipulation policies conditioned on vision and language~\cite{brohan2022rt,brohan2023rt,team2024octo,kim2024openvla}. Follow-up work has explored continuous action generation and efficient deployment~\cite{black2024pi_0,li2024cogact,shukor2025smolvla}. To improve planning on complex tasks, other approaches integrate intermediate reasoning into action generation through embodied chain-of-thought~\cite{zawalski2024robotic}, visual subgoals~\cite{zhao2025cot}, and latent planning~\cite{huang2026thinkact}. These approaches strengthen policies through model architecture and training design, while adaptation at deployment addresses failures on tasks that pretraining data do not adequately cover.

\textbf{Test-time adaptation for VLA models.} Tent and CoTTA established approaches for online and continual adaptation, respectively~\cite{wang2020tent,wang2022continual}. TPT and TDA extended TTA to vision--language models through prompt updates and dynamic caching, respectively~\cite{shu2022test,karmanov2024efficient}. In embodied settings, fast and slow updates and feedback-driven reinforcement learning have been used for online vision--language navigation~\cite{gao2023fast,kim2025test}. TT-VLA, VANE, and PDF use task progress, future visual observations, and delayed feedback, respectively, to adapt VLA models~\cite{liu2026fly,ji2026vane,zang2026test}. Retrieve-then-Steer guides action generation by retrieving successful past experiences~\cite{zhao2026retrieve}. For methods that fine-tune the model at test time, limited on-device resources constrain the updates they can perform.

\textbf{Optimization methods for test-time adaptation.} FOA, FOZO, and CAZO reduce the cost of TTA by avoiding backpropagation through evolutionary strategies, ZO prompt optimization, and curvature-aware perturbations, respectively~\cite{niu2024test,wang2026fozo,zhang2026curvature}. ZO-SVRG, MeZO-SVRG, and ZO-AdaMM explore variance reduction and the use of historical gradient information~\cite{gautam2024variance,chen2019zo,liu2018zeroth}. MeZO and Dust apply ZO optimization to language model fine-tuning and Transformer pretraining, respectively~\cite{malladi2023fine,dahal2026backprop}. In long-horizon VLA-TTA, perturbation noise can overwhelm the dominant error signal and weaken parameter updates. This motivates Dune's approach to the performance plateau problem: combining current ZO perturbation observations with regression predictions from historical statistics to obtain more reliable estimates of layerwise error signals.
\section{Motivation}
\label{sec:motivation}

The preceding results demonstrate the memory advantage and early convergence potential of ZO optimization for VLA-TTA. Maintaining stable updates over long horizons of online adaptation remains an open challenge. Two questions guide the design of \textbf{Dune}:

\begin{itemize}
    \item Why does ZO optimization degrade over long adaptation horizons in VLA-TTA?
    \item How can we mitigate this degradation without backpropagation?
\end{itemize}

The mean squared error of a ZO gradient estimate depends strongly on the dimensionality of the model parameters. Lemma~\ref{lem:gaussian-gradient-mse} quantifies this dependence for a Gaussian ZO gradient estimator.

\begin{lemma}[MSE of ZO Gradient Estimation~\cite{berahas2021theoreticalempiricalcomparisongradient}]
\label{lem:gaussian-gradient-mse}
Let $x\in\mathbb{R}^d$ denote the model parameters, where $d$ is the parameter dimension. Consider the linear objective $f(x)=a^\top x$, with $a\in\mathbb{R}^d\setminus\{0\}$. Given a perturbation scale $\sigma>0$ and a perturbation population size $K\in\mathbb{N}_{+}$, the Gaussian ZO gradient estimator $\widehat{g}_K(x)$ constructed from independent and identically distributed Gaussian perturbations $u_1,\ldots,u_K \overset{\mathrm{i.i.d.}}{\sim}\mathcal{N}(0,I_d)$ satisfies:
\begin{equation}
    \mathbb{E}\bigl[
        \|\widehat{g}_K(x)-\nabla f(x)\|^2
    \bigr]
    =
    \frac{d+1}{K}\|a\|^2.
\end{equation}
\end{lemma}

Lemma~\ref{lem:gaussian-gradient-mse} shows that the variance of the Gaussian ZO gradient estimate scales linearly with the parameter dimension $d$ and inversely with the perturbation population size $K$. During adaptation in nonlinear deep networks, the norm of the true gradient, which provides the dominant signal for the update, decreases as the model approaches a locally optimal region. On-device memory and compute budgets typically constrain the perturbation population to be much smaller than the parameter dimension ($d\gg K$). In this regime, sampling variance can account for a growing share of the update as the true gradient weakens. The resulting decline in the \textbf{signal-to-noise ratio (SNR)} allows perturbation noise to gradually overwhelm the dominant signal. To quantify how estimation variance affects long-horizon convergence, we compare the loss recursions of first-order gradient descent (FO-GD) and ZO optimization on a convex quadratic objective in Theorem~\ref{thm:fo-zo-loss-recursion}.

\begin{assumption}[Convex quadratic objective]
\label{assumption1}
Let $x\in\mathbb{R}^d$ denote the model parameters. The objective has the quadratic form:
\begin{equation}
    f(x)=\frac{1}{2}(x-x^\star)^\top H(x-x^\star),
    \quad x\in\mathbb{R}^d,
\end{equation}
where $x^\star$ is the global minimizer and $H\succ 0\in\mathbb{R}^{d\times d}$.
\end{assumption}

\begin{theorem}[Loss recursions for zeroth-order optimization and first-order gradient descent]
\label{thm:fo-zo-loss-recursion}
Under Assumption~\ref{assumption1}, let $g(x)=\nabla f(x)$ denote the true gradient. The parameter updates for FO-GD and ZO optimization are, respectively,
\begin{equation}
    \begin{aligned}
        &\textbf{FO-GD:}\quad
        x^g_{t+1}=x^g_t-\eta_t^g g(x_t^g),\\
        &\textbf{ZO:}\quad
        x^z_{t+1}=x^z_t-\eta_t^z \widehat{g}_K(x_t^z),
    \end{aligned}
\end{equation}
where $\eta_t^g,\eta_t^z>0$ are the corresponding learning rates. The one-step loss recursions are:
\begin{equation}
    \begin{aligned}
        f(x_{t+1}^g)
        &=
        f(x_t^g)
        -
        \eta_t^g\|g(x_t^g)\|^2
        +
        \frac{(\eta_t^g)^2}{2}
        g(x_t^g)^\top H g(x_t^g),\\
        \mathbb{E}_t[f(x_{t+1}^z)]
        &=
        f(x_t^z)
        -
        \eta_t^z\|g(x_t^z)\|^2
        +
        \frac{(\eta_t^z)^2}{2}
        g(x_t^z)^\top H g(x_t^z)\\
        &\quad+
        \underbrace{
            \frac{(\eta_t^z)^2}{2K}
            \left[
                g(x_t^z)^\top H g(x_t^z)
                +
                \operatorname{tr}(H)\|g(x_t^z)\|^2
            \right]
        }_{\text{ZO estimation variance term}}.
    \end{aligned}
\end{equation}
In the isotropic case ($H=\mu I_d$, $\mu>0$), if each method uses its optimal learning rate at every step, the losses after $T>2$ iterations satisfy:
\begin{equation}
    \begin{aligned}
        &\textbf{FO-GD:}\quad
        f(x_T^g)=0,\\
        &\textbf{ZO:}\quad
        \mathbb{E}[f(x_T^z)]
        =
        \left(
            \frac{d+1}{K+d+1}
        \right)^T f(x_0).
    \end{aligned}
\end{equation}
The proof of Theorem~\ref{thm:fo-zo-loss-recursion}
refers to Appendix.
\end{theorem}

\noindent\textbf{Remark 1:} Theorem~\ref{thm:fo-zo-loss-recursion} shows that FO-GD rapidly drives the loss to zero with the optimal step size, whereas the convergence of ZO optimization is governed by the dimension-dependent factor $\frac{d+1}{K+d+1}$. In online VLA adaptation, the parameter dimension $d$ is large, while inference latency and memory budgets keep the per-step perturbation population size $K$ small. When $d\gg K$, the factor $\frac{d+1}{K+d+1}$ is close to one, so the additional variance term limits how much the loss can decrease within a finite TTA budget of $T$ steps. Combined with the accumulation of nonstationary noise under online distribution shifts, this slowdown helps explain the \textbf{performance plateau problem} observed in long-horizon VLA-TTA.

When the deployment budget limits $K$, we turn to historical observations to recover the dominant signal. During VLA-TTA on a continuous video stream, neighboring observations are closely related in space and time, and model parameters change gradually. Local error signals can therefore share structure across batches. Simply averaging past update directions gives limited ability to track changes in current output residuals and flow-matching timesteps. \textbf{Dune} instead uses \textbf{conditional historical regression}. It conditions on output residuals and flow-matching time buckets, uses historical decay and ridge regularization to fit a mapping from these features to each layer's local error signal online, and forms a weighted combination of the historical regression predictions and the current ZO perturbation observations. This mechanism reduces estimation variance without additional forward evaluations for perturbations, yielding a higher-SNR update signal for long-horizon adaptation.
\section{Method}
\label{sec:method}

\begin{figure*}[t]
    \centering
    \includegraphics[width=\textwidth]{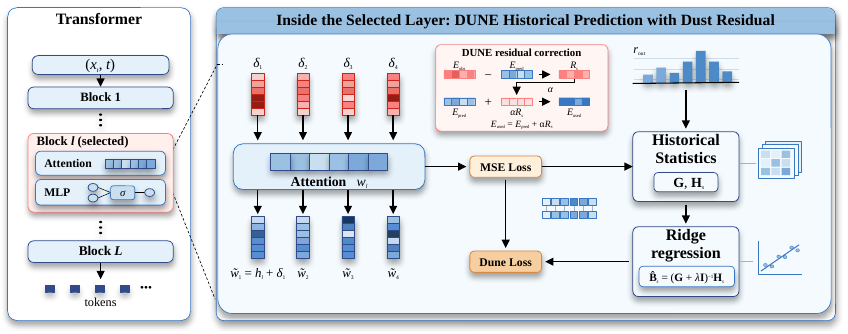}
    \caption{Overview of Dune. The left panel shows the target layers in the Transformer, and the right panel illustrates activation perturbation, conditional historical regression, and error fusion. The upper inset depicts the residual between the current zeroth-order perturbation observation and the historical regression prediction, and shows how weighted fusion corrects the prediction.}
    \label{fig:dune-overview}
\end{figure*}

We propose \textbf{Dune}, which extends Dust's activation perturbation estimator with conditional historical regression. Dune combines historical regression predictions with current zeroth-order (ZO) perturbation observations to update model parameters. Figure~\ref{fig:dune-overview} summarizes the method. During adaptation, Dune freezes the vision-language backbone and fine-tunes only the action expert and its associated projection layers. The procedure requires no global backpropagation.

\subsection{Dust Activation Perturbation Estimation}
\label{subsec:dust}

Given the vision-language condition $c$ and target action $a$, we sample Gaussian noise $\epsilon$ and a flow-matching timestep $\tau\in[0,1]$ to construct the interpolated state $x_\tau=(1-\tau)a+\tau\epsilon$. With the target velocity field $u=\epsilon-a$, we minimize the mean squared error (MSE) between the model prediction $v=f_\theta(c,x_\tau,\tau)$ and $u$.
\begin{algorithm}[ht]
\caption{Dune}
\label{alg:DecFus}
\KwIn{Dataset $\mathcal{D}$, initial model $\theta^0$,
target layers $\mathcal{S}$, adaptation steps $T$,
warm-up steps $T_{\mathrm w}$, and hyperparameters
$B,\{\sigma_s,K_s\}_{s\in\mathcal{S}},\rho,\beta,\alpha$}
\KwOut{Adapted model $\theta^T$}

\textbf{Initialization:}
Freeze the vision-language backbone;
set $G_0=H_{s,0}=0$ for all $s\in\mathcal{S}$\;

\For{$k \in \{1,\ldots,T\}$}{
    Sample a mini-batch from $\mathcal{D}$,
    Gaussian noise $\epsilon$, and timesteps $\tau$\;
    \If{$k=1$}{
        Define $B$ fixed time buckets using
        first-batch timestep quantiles\;
    }
    Run an unperturbed forward pass and cache activations\;

    \tcp{Activation perturbation estimation}
    \For{$s \in \mathcal{S}$}{
        $\widehat E_{s,k}^{\mathrm D}
        \leftarrow \text{Dust perturbation estimation}$
        by (\ref{eq:dust_teacher})\;
    }

    \tcp{Time-conditioned error prediction}
    Construct $\Phi_k$ by (\ref{eq:feedback_prediction})\;
    \For{$s \in \mathcal{S}$}{
        $E_{s,k}^{\mathrm F} \leftarrow 0$\;
        \If{$k>T_{\mathrm w}$ \textbf{and} $G_{k-1}\neq 0$}{
            Solve for $C_{s,k}$ using historical statistics
            by (\ref{eq:ridge_solve})\;
            Compute $E_{s,k}^{\mathrm F}$
            by (\ref{eq:feedback_prediction})\;
        }
    }

    \tcp{Historical regression and fusion updates}
    \For{$s \in \mathcal{S}$}{
        Form $E_{s,k}^{\mathrm U}$ with warm-up handling
        by (\ref{eq:error_fusion})\;
    }
    Update $G_k$ and $\{H_{s,k}\}_{s\in\mathcal{S}}$
    using raw ZO observations by (\ref{eq:history_update})\;
    Update trainable parameters using
    $\{E_{s,k}^{\mathrm U}\}_{s\in\mathcal{S}}$
    to obtain $\theta^k$\;
}
\Return{$\theta^T$}\;
\end{algorithm}

Following Dust~\cite{dahal2026backprop}, we first run an unperturbed forward pass and cache the activations at each layer. We then inject a Gaussian perturbation $\delta_s\xi_s^{(i)}$ into target layer $s$, where $\delta_s=\sigma_s\operatorname{RMS}(h_s^{(0)})$ is the perturbation scale and $\xi_s^{(i)}$ is standard Gaussian noise. We compute the within-group centered reward $\widetilde c_{bt}^{(i)}$ from the reduction in loss under perturbation and obtain the current ZO perturbation observation as:
\begin{equation}
    \widehat e_{s,bt}^{\mathrm D}
    =-\frac{1}{K_s\delta_s}
    \sum_{i=1}^{K_s}
    \widetilde c_{bt}^{(i)}\xi_{s,bt}^{(i)},
    \label{eq:dust_teacher}
\end{equation}
where $(b,t)$ indexes samples and action tokens, and $K_s$ is the perturbation population size. Perturbation evaluation reuses cached upstream activations and recomputes only the affected downstream layers. Perturbations to the Q/K/V projections are scored using the local error at the attention output. For the action output head, we perturb the cached predictions directly and evaluate the resulting loss.

\subsection{Time-Conditioned Error Prediction}
\label{subsec:conditional_feedback}

Dune uses historical perturbation observations to learn an online mapping from model output residuals to local error signals at each target layer. Let $r_{bt}=2M_{bt}(v_{bt}-u_{bt})$ denote the output residual, where $M_{bt}$ is the valid-token mask. We use quantiles of the flow-matching timesteps in the first batch to define $B$ fixed flow-matching time buckets that capture differences in dynamics across denoising stages. We define the conditional regression features and error predictions as:
\begin{equation}
    \phi_{bt}
    =\operatorname{onehot}(q(\tau_b))\otimes r_{bt},
    \qquad
    e_{s,bt}^{\mathrm F}=C_s^\top\phi_{bt},
    \label{eq:feedback_prediction}
\end{equation}
where $q(\tau_b)$ is the flow-matching time bucket index for sample $b$ and $C_s\in\mathbb{R}^{p\times d_s}$ is the mapping matrix. Here, $p=Bm$, $m$ is the action dimension, and $d_s$ is the activation width of layer $s$. This construction gives each time bucket its own residual-to-error mapping.

For batch $k$, we stack the conditional features and ZO perturbation observations row-wise into $\Phi_k$ and $\widehat E_{s,k}^{\mathrm D}$, respectively. The batched historical regression prediction is then $E_{s,k}^{\mathrm F}=\Phi_kC_{s,k}$. Dune uses $\widehat E_{s,k}^{\mathrm D}$ directly as the regression target, so fitting the mapping requires no explicit computation of the network Jacobian.

\subsection{Historical Regression and Fusion Updates}
\label{subsec:online_ridge}

Dune estimates the mapping matrix online using ridge regression with historical decay. It maintains shared feature statistics $G$ and layer-specific cross-statistics $H_s$. For the current batch, it solves for the mapping using statistics accumulated through the previous batch:
\begin{equation}
    (G_{k-1}+\lambda_{k-1}I)C_{s,k}=H_{s,k-1},
    \lambda_{k-1}
    =\rho\frac{\operatorname{tr}(G_{k-1})}{p},
    \label{eq:ridge_solve}
\end{equation}
where $\rho>0$ controls the regularization strength. If the historical statistics are zero, Dune sets the historical regression prediction to zero and skips the solve.

During the first $T_{\mathrm w}$ adaptation steps, Dune updates the parameters using only the ZO perturbation observations while accumulating historical statistics. After this warm-up period, it combines the historical regression prediction with the current ZO perturbation observation:
\begin{equation}
    E_{s,k}^{\mathrm U}
    =
    \begin{cases}
        \widehat E_{s,k}^{\mathrm D},
        & k\leq T_{\mathrm w},\\
        (1-\alpha)E_{s,k}^{\mathrm F}
        +\alpha\widehat E_{s,k}^{\mathrm D},
        & k>T_{\mathrm w}.
    \end{cases}
    \label{eq:error_fusion}
\end{equation}
The historical regression prediction $E_{s,k}^{\mathrm F}$ captures the structure of the dominant error signal across batches, while the current ZO perturbation observation $\widehat E_{s,k}^{\mathrm D}$ keeps the update responsive to the latest model state. After forming the fused error for the current batch, Dune updates the historical statistics online:
\begin{equation}
    \begin{aligned}
        G_k
        &=\beta G_{k-1}+\Phi_k^\top\Phi_k,\\
        H_{s,k}
        &=\beta H_{s,k-1}
          +\Phi_k^\top\widehat E_{s,k}^{\mathrm D}.
    \end{aligned}
    \label{eq:history_update}
\end{equation}
We initialize $G_0=H_{s,0}=0$. The historical decay coefficient $\beta$ progressively downweights older observations. The regression always uses raw ZO perturbation observations as targets, and the statistics can be updated without storing previous samples.

Building on the three main components described above, we detail Dune in Algorithm~\ref{alg:DecFus}.

\section{Experiments}
\label{sec:exp}
To verify the effectiveness of Dune for VLA-TTA, we will examine it from the following four perspectives:
\begin{itemize}
    \item[1.] \textbf{Adaptation Performance:} \textit{Can Dune effectively fine-tune VLA models and overcome performance plateaus during long-horizon test-time adaptation?} In Section~5.2, we evaluate Dune on the LIBERO simulation benchmark and real-world robotic platforms against FO and zeroth-order ZO baselines.
\item[2.] \textbf{Ablation Study:} \textit{How does time-conditioned historical regression improve adaptation?} In Section~5.3, we compare policies fine-tuned with and without time-conditioned historical regression to examine its effect on closed-loop trajectory stability and task recovery under external disturbances.    
\item[3.] \textbf{Computational Cost:} \textit{Does Dune maintain a low memory footprint during adaptation?} In Section~5.4, we measure the peak memory consumption and runtime overhead of Dune, demonstrating substantial memory savings over backpropagation-based methods with negligible overhead relative to pure zeroth-order baselines.
    \item[4.] \textbf{Interpretability:} \textit{Does Dune exhibit favorable optimization dynamics?} In Section~5.5, we inspect the curvature structures of Adam, Dust, and Dune throughout training to understand how historical error prediction preserves block-diagonal update patterns.
\end{itemize}
\begin{table}[t]
    \centering
    \caption{Task success rates (SR, \%) on LIBERO after 24{,}000 steps of fine-tuning from the pretrained SmolVLA checkpoint. For Dust, we report task success rates at step 12{,}000, during the performance plateau. All zeroth-order methods use a perturbation population size of $K=8$. Bold values indicate the \textbf{best reported results}. Underlined values indicate the \underline{second-best reported results}.}
    \label{tab:libero_results}
    \renewcommand{\arraystretch}{1.2}
    \small
    \setlength{\tabcolsep}{4.8pt}
    \begin{tabular}{@{}c ccccc c@{}}
        \toprule
        Method & Type & Spatial & Object & Goal & Long & Mean SR (\%) \\
        \midrule
        SGD & FO & 24.0 & 34.0 & 31.6 & 7.2 & 24.2 \\
        SignSGD & FO & 62.0 & 58.2 & 63.0 & 24.8 & 52.0 \\
        AdamW & FO & \textbf{74.6} & \textbf{85.4} & \textbf{88.8} & \textbf{64.4} & \textbf{78.3} \\
        \midrule
        MeZO & ZO & 0.0 & 0.0 & 0.0 & 0.0 & 0.0 \\
        n-SPSA & ZO & 0.0 & 0.0 & 0.0 & 0.0 & 0.0 \\
        Dust & ZO & 46.4 & 59.4 & 57.2 & 26.8 & 47.5 \\
        \textbf{Dune} & ZO & \underline{64.2} & \underline{77.2} & \underline{69.0} & \underline{35.2} & \underline{61.4} \\
        \bottomrule
    \end{tabular}
\end{table}
\begin{table}[t]
    \centering
    \caption{Task success rate (SR) and mean completion time after fine-tuning on the real-world bimanual task with the AgileX PiPER-X robot.}
    \label{tab:real_world_results}
    \renewcommand{\arraystretch}{1.2}
    \small
    \setlength{\tabcolsep}{8pt}
    \begin{tabular}{@{}ccc@{}}
        \toprule
        Method & SR (\%) & Mean completion time (s) \\
        \midrule
        AdamW & 32.0 & 102 \\
        Dune & 26.0 & 144 \\
        \bottomrule
    \end{tabular}
\end{table}
\begin{figure*}[t]
  \centering
  \setlength{\tabcolsep}{0pt}
  \makebox[\linewidth][l]{\small\textbf{(a) LIBERO simulation}}\par\vspace{3pt}
  \begin{tabular*}{\linewidth}{@{\extracolsep{\fill}}cccc@{}}
    \includegraphics[width=0.23\linewidth]{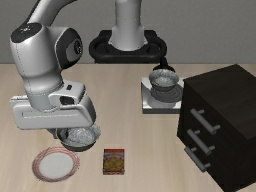} &
    \includegraphics[width=0.23\linewidth]{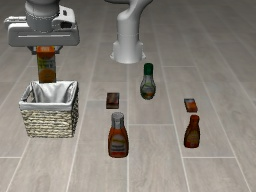} &
    \includegraphics[width=0.23\linewidth]{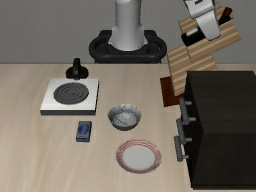} &
    \includegraphics[width=0.23\linewidth]{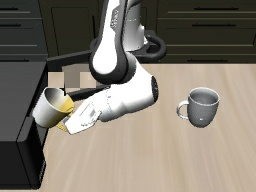} \\[2pt]
    \small Spatial (bowl to plate) & \small Object (juice to basket) &
    \small Goal (bottle to rack) & \small Long (mug into microwave)
  \end{tabular*}\par\vspace{7pt}
  \makebox[\linewidth][l]{\small\textbf{(b) PiperX real-world demonstration}}\par\vspace{3pt}
  \begin{tabular*}{\linewidth}{@{\extracolsep{\fill}}ccccccc@{}}
    \includegraphics[width=0.23\linewidth]{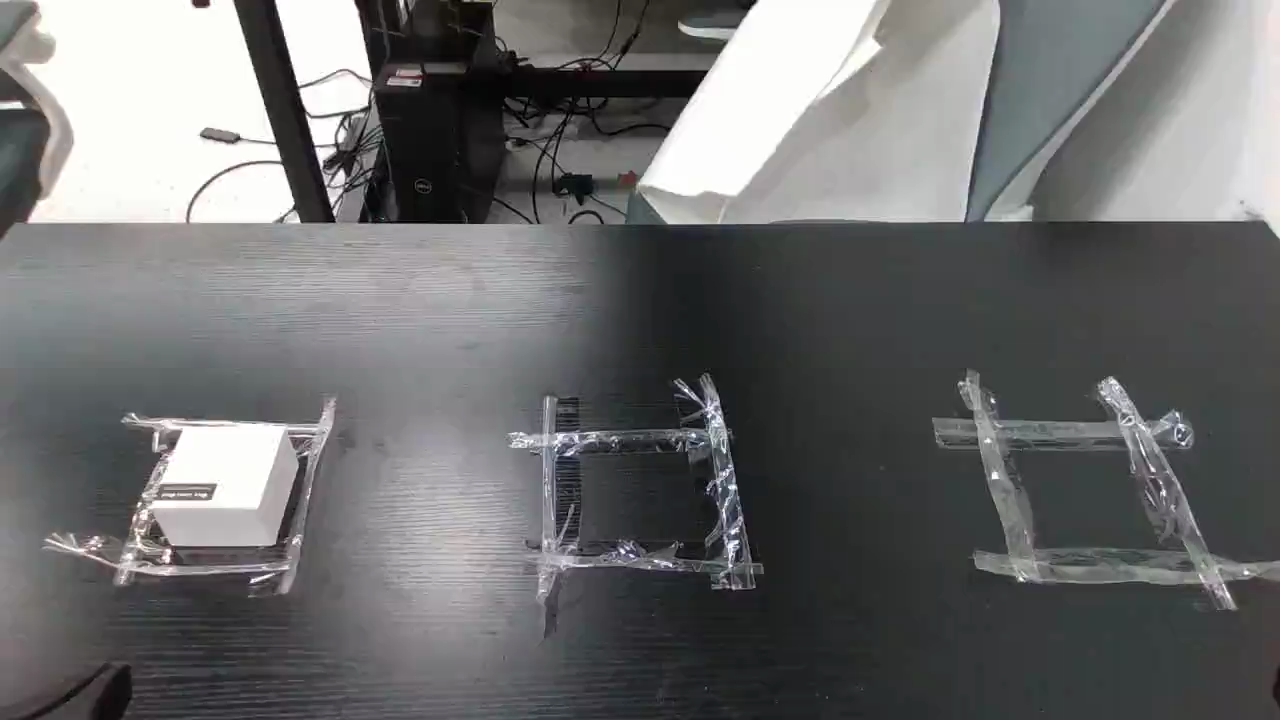} &&
    \includegraphics[width=0.23\linewidth]{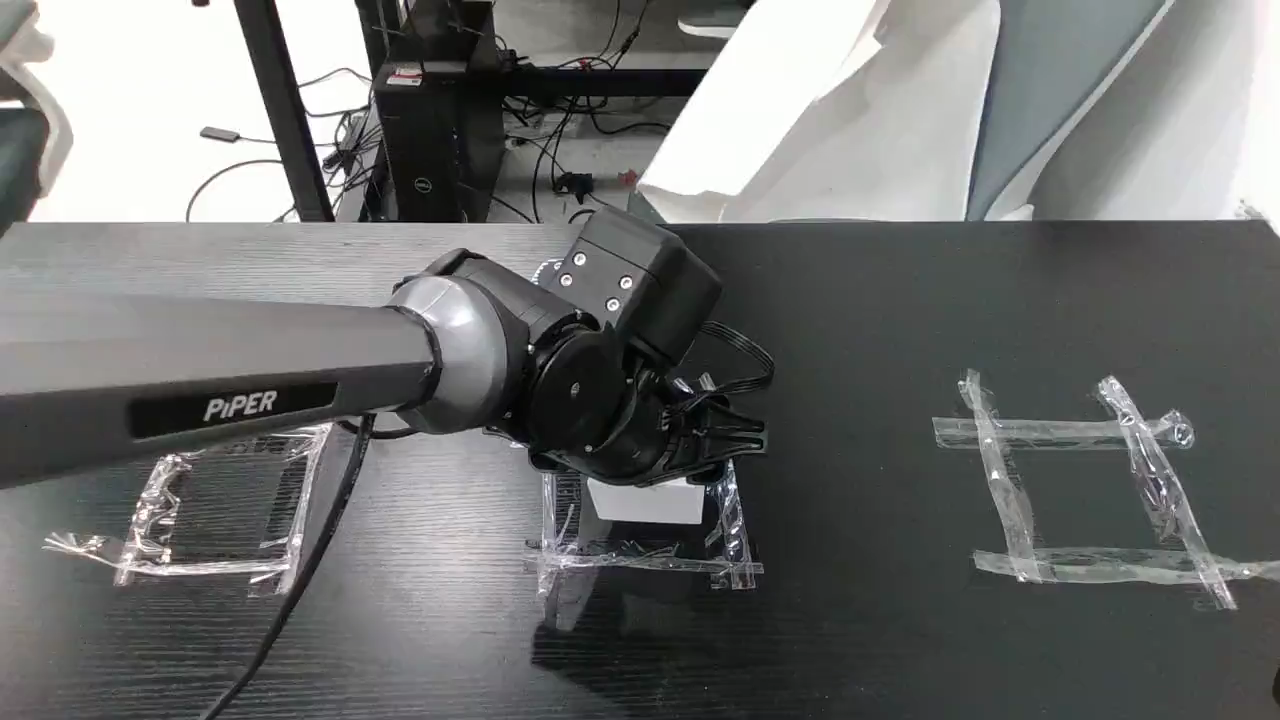} &&
    \includegraphics[width=0.23\linewidth]{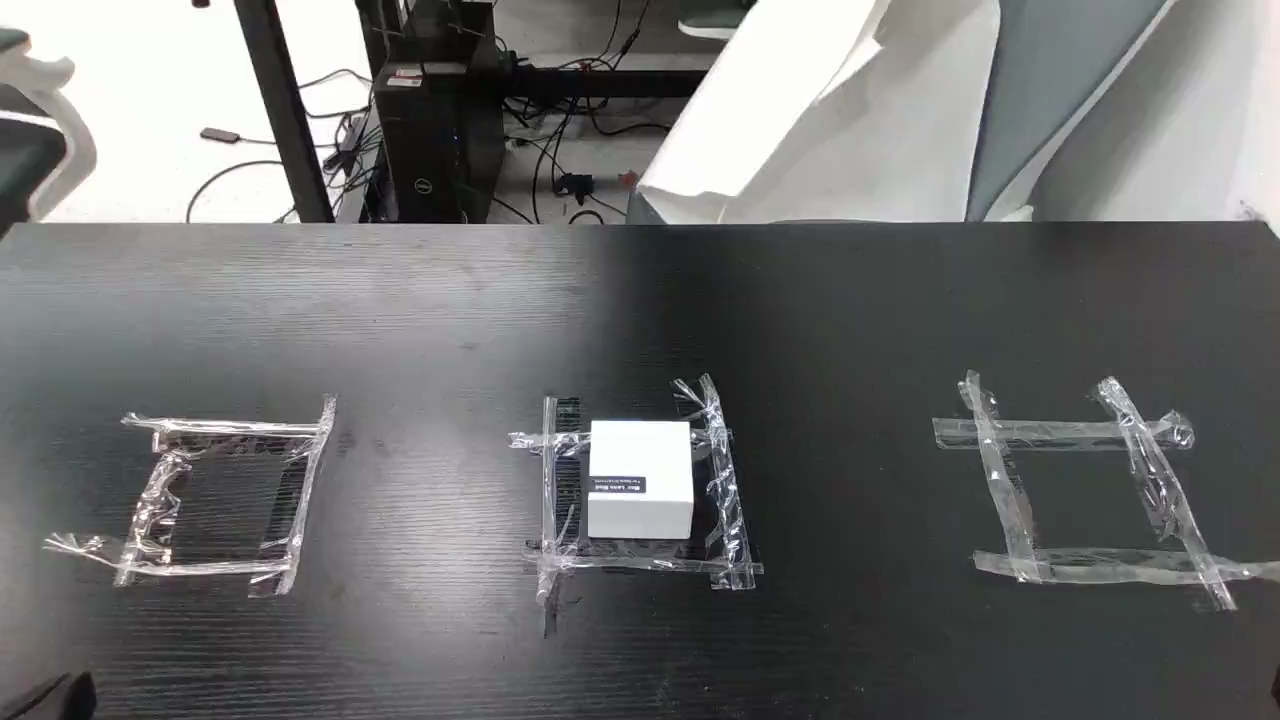} &&
    \includegraphics[width=0.23\linewidth]{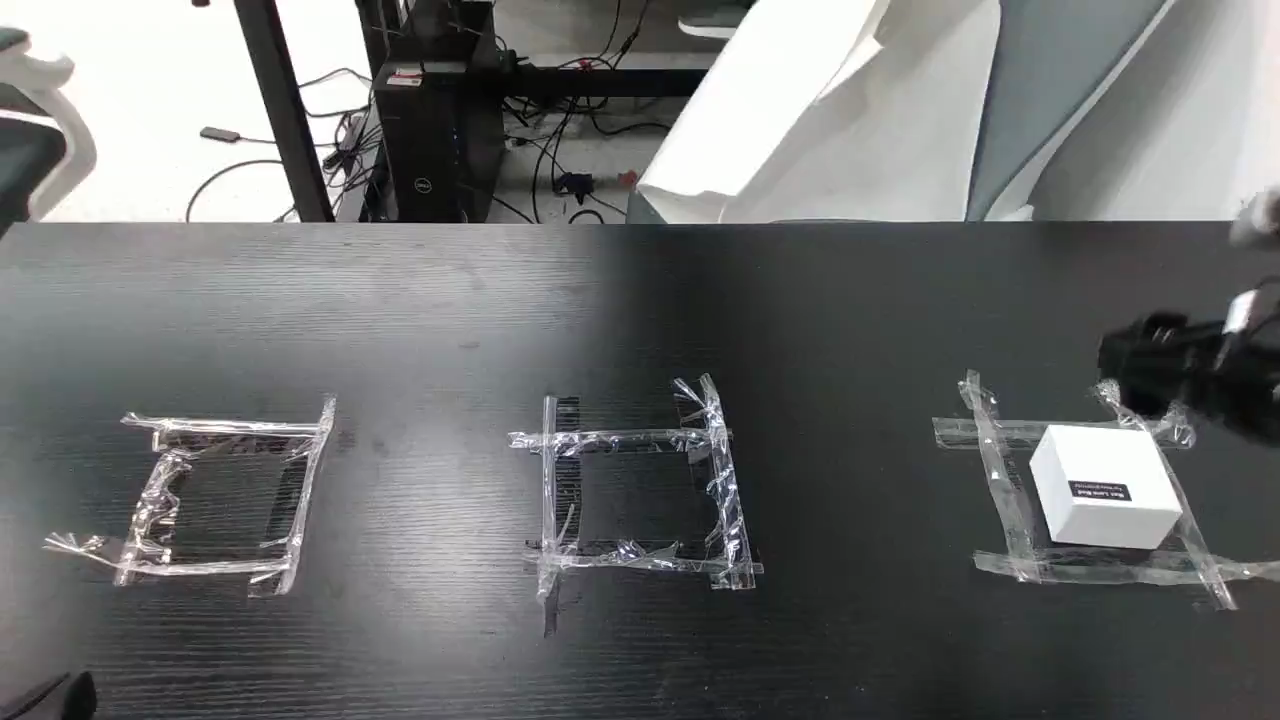} \\[2pt]
    \small Start zone & \(\rightarrow\) &
    \small Left-arm transfer & \(\rightarrow\) &
    \small Handoff zone & \(\rightarrow\) &
    \small Right-arm placement
  \end{tabular*}
  \caption{Simulation and real-world task settings. (a) Workspace crops from the external camera show one training demonstration per LIBERO suite. (b) A recorded PiperX box transfer: the left arm moves the box to the center handoff zone, and the right arm places it in the goal zone.}
  \label{fig:experimental-settings}
\end{figure*}
\begin{figure*}[t]
    \centering
    \includegraphics[width=\textwidth]{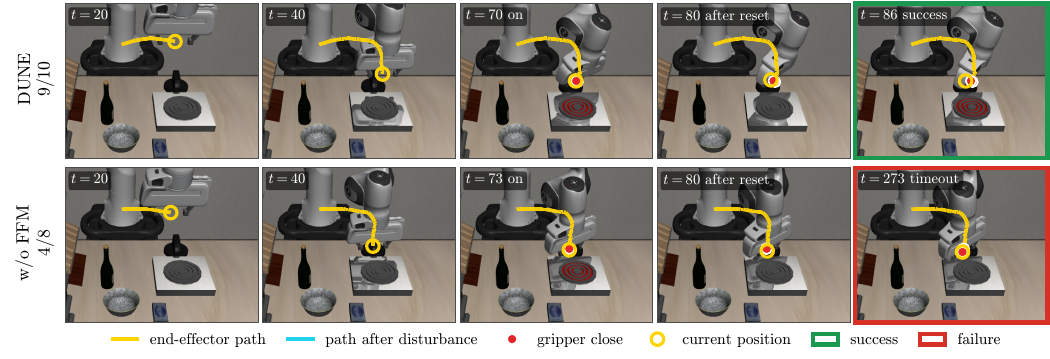}
    \caption{Recovery from an external disturbance in the stove task (initial state 1), using the same policies fine-tuned for 12{,}000 steps as in Figure~\ref{fig:fan}. After each policy first completes the task, we manually reset the knob at $t=70$ for \textbf{Dune} and at $t=73$ for the policy without historical regression. \textbf{Dune} adjusts its actions and turns the stove back on at $t=86$. The policy without historical regression stalls at the knob until the 200-step recovery budget is exhausted. In the evaluation covering 10 initial states, the reported recovery counts are $9/10$ for \textbf{Dune} and $4/8$ for the policy without historical regression.}
    \label{fig:recover}
\end{figure*}

\subsection{Experimental Setting}
\label{sec:setup}
\textbf{Model and simulation benchmark.}
In the main experiments, we fine-tune SmolVLA, a 240M-parameter VLA model, during test-time adaptation. We freeze the SmolVLM2 vision-language backbone and update only the action expert and its associated projection layers. We train and evaluate on all four task suites of the \textbf{LIBERO} benchmark, which comprise 40 manipulation tasks, 1{,}693 expert demonstration trajectories, and 273{,}465 image frames. At each timestep, the model receives $256\times256$ RGB images from two viewpoints, the robot's proprioceptive state, and a natural language instruction. Figure~\ref{fig:experimental-settings} shows representative simulation scenes from the four suites. We use task success rate (SR) as the primary metric across all benchmarks. In LIBERO, each rollout receives a score of 1 if the task succeeds and 0 otherwise.

\textbf{Real-world robot task.}
We evaluate deployment on an AgileX PiPER-X bimanual robot equipped with three RGB cameras. An inference node with a single NVIDIA GeForce RTX 4090 GPU runs the policy at a target control frequency of 40\,Hz. We design a bimanual box transfer task in which the robot alternates between its left and right arms to move a box from the left side of the workbench, through an intermediate handoff point, to a target area on the right. Figure~\ref{fig:experimental-settings}  shows the three interaction locations marked with tape.

\textbf{Baselines.}
To evaluate \textbf{Dune}, we compare three FO optimizers and three ZO optimizers that operate without backpropagation. We fine-tune SmolVLA with each optimizer under the same test-time adaptation protocol.
\begin{itemize}
    \item \textbf{SGD (FO).} Standard stochastic gradient descent updates parameters using FO gradients computed by backpropagation.
    \item \textbf{SignSGD (FO).} This sign-based optimizer uses only the signs of the backpropagated gradients to take fixed-size steps. Sign-based updates are commonly used to reduce the cost of communicating and quantizing gradients.
    \item \textbf{AdamW (FO).} This adaptive optimizer uses decoupled weight decay and serves as the performance reference for backpropagation-based fine-tuning.
    \item \textbf{n-SPSA (ZO).} This multi-sample simultaneous perturbation stochastic approximation method estimates gradients from forward-pass loss differences across multiple random perturbations in parameter space.
    \item \textbf{MeZO (ZO).} This memory-efficient optimizer uses in-place random perturbations and two forward passes to estimate gradients.
    \item \textbf{Dust (ZO).} This state-of-the-art optimizer uses activation-space perturbations to estimate error signals.
\end{itemize}

\subsection{Evaluation of Dune}
\textbf{Simulation results.}
Table~\ref{tab:libero_results} reports task success rates after 24{,}000 steps of test-time fine-tuning from the pretrained SmolVLA checkpoint. MeZO and n-SPSA achieve a 0\% success rate in every suite. Dust makes initial progress with activation-space perturbations, but its performance degrades over long adaptation horizons. \textbf{Dune} achieves a mean success rate of 61.4\% without global backpropagation. It outperforms SGD (24.2\%) and SignSGD (52.0\%) on all four suites and narrows the gap to AdamW (78.3\%). These results support the use of conditional historical regression to suppress estimation variance in high-dimensional settings.

\textbf{Real-world results.}
We collect 99 teleoperated demonstration trajectories on the physical robot, totaling 90{,}904 image frames, to fine-tune SmolVLA. A trial succeeds if the robot completes the bimanual box transfer within 200\,s. As shown in Table~\ref{tab:real_world_results}, \textbf{Dune} achieves a 26\% success rate without global backpropagation, within 6 percentage points of AdamW (32\%). This result supports the feasibility of deploying Dune on a physical robot.

\begin{figure}[htbp]
\centering
\begin{subfigure}[t]{0.265\linewidth}
    \centering
\includegraphics[width=\linewidth]{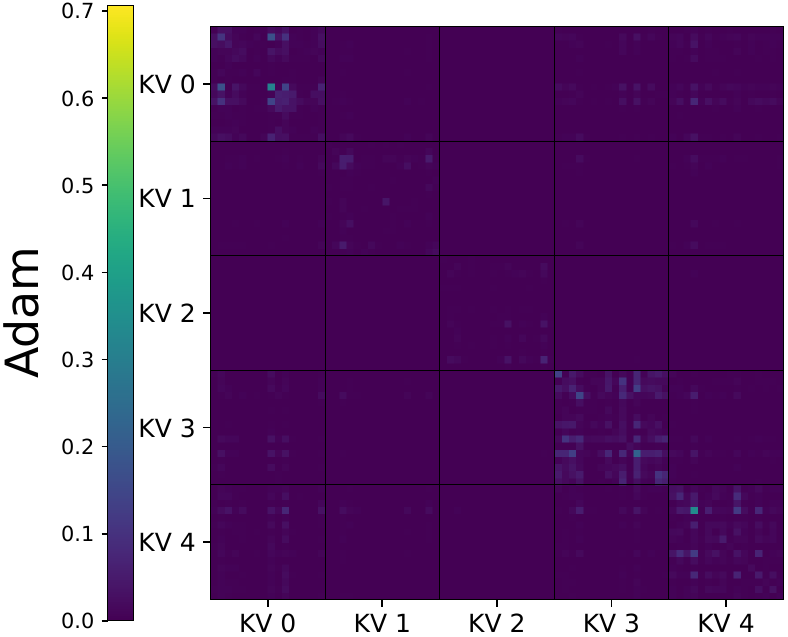}
    \caption{2000 Steps}
    \label{fig:ex2-hyperball-to-weight-decay}
\end{subfigure}
\begin{subfigure}[t]{0.22\linewidth}
    \centering
\includegraphics[width=\linewidth]{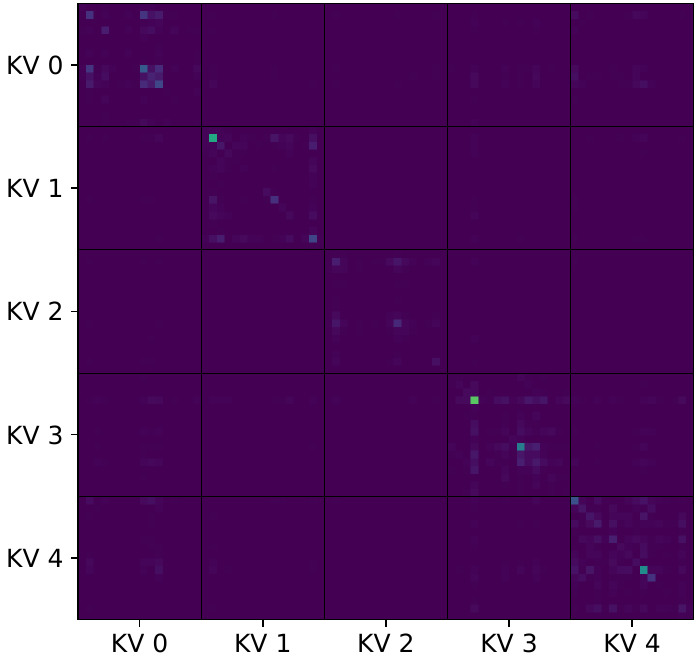}
    \caption{8000 Steps}
    \label{fig:ex2-hyperball-to-weight-decay}
\end{subfigure}
\begin{subfigure}[t]{0.22\linewidth}
    \centering
\includegraphics[width=\linewidth]{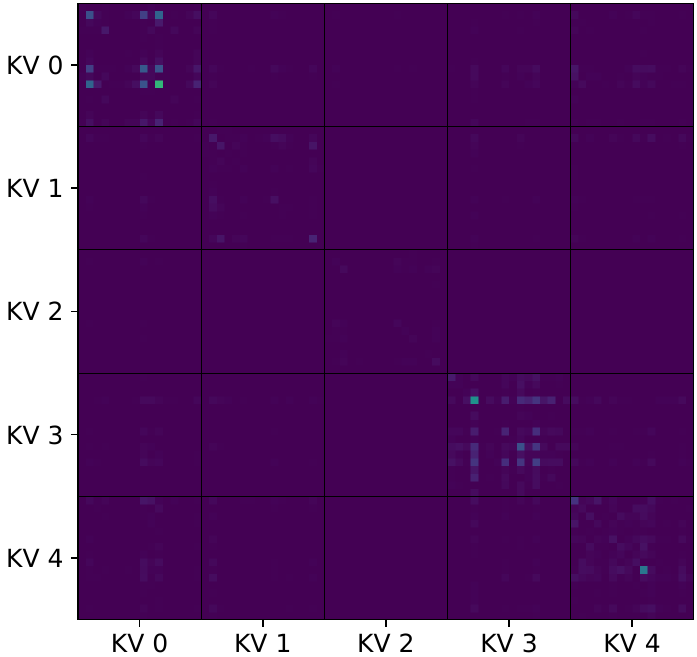}
    \caption{12000 Steps}
    \label{fig:ex2-hyperball-to-weight-decay}
\end{subfigure}
\begin{subfigure}[t]{0.22\linewidth}
    \centering
\includegraphics[width=\linewidth]{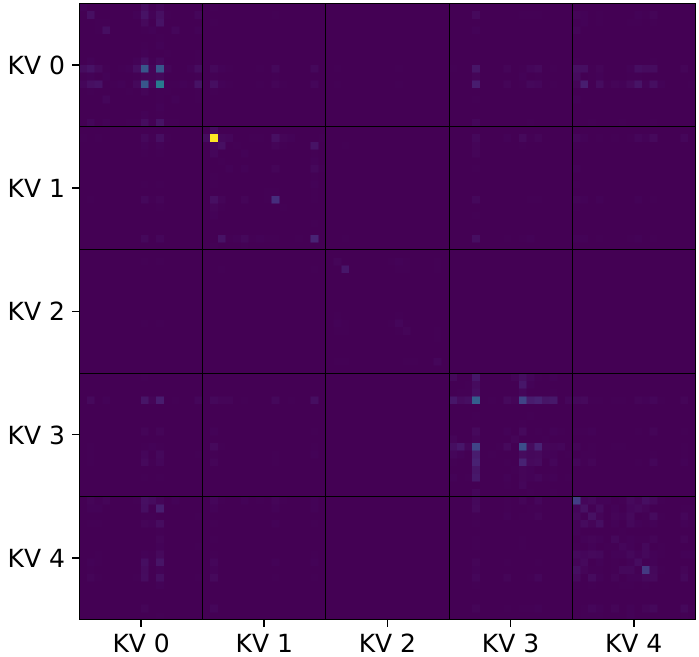}
    \caption{18000 Steps}
    \label{fig:ex2-hyperball-to-weight-decay}
\end{subfigure}
\begin{subfigure}[t]{0.265\linewidth}
    \centering
\includegraphics[width=\linewidth]{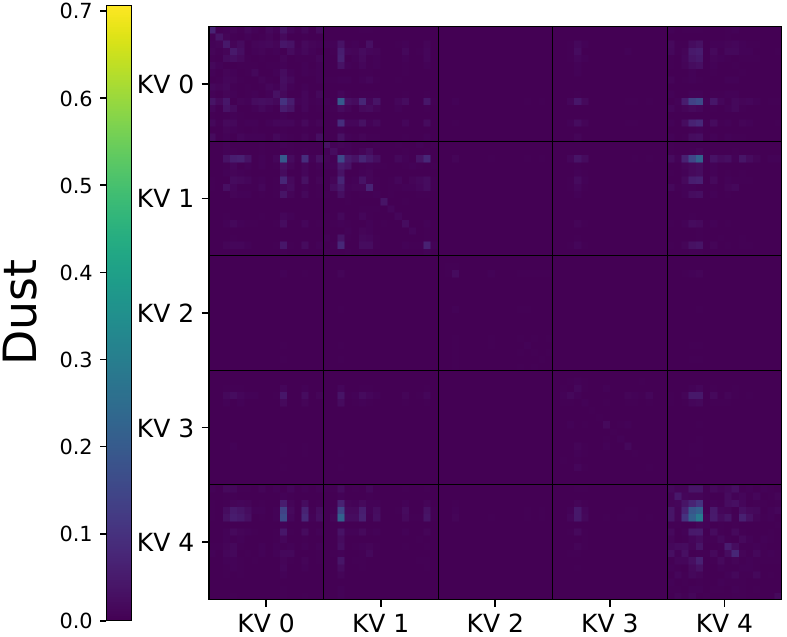}
    \caption{2000 Steps}
    \label{fig:ex2-hyperball-to-weight-decay}
\end{subfigure}
\begin{subfigure}[t]{0.22\linewidth}
    \centering
\includegraphics[width=\linewidth]{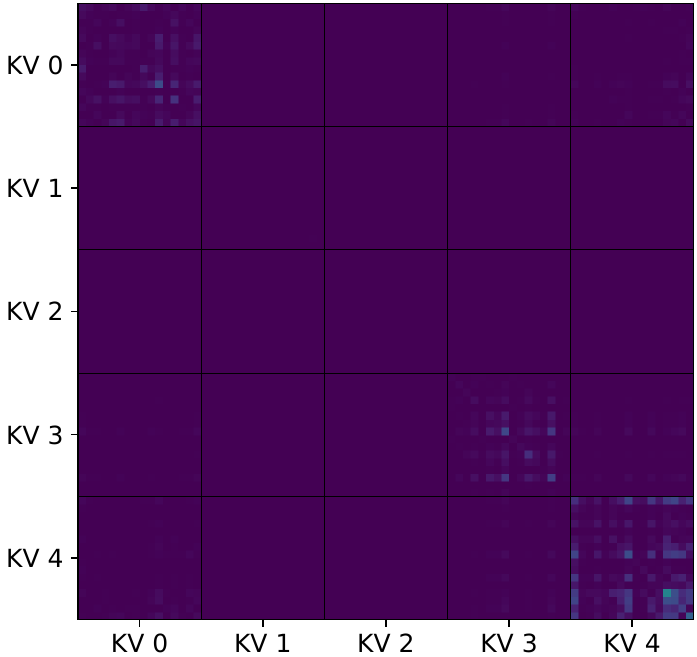}
    \caption{8000 Steps}
    \label{fig:ex2-hyperball-to-weight-decay}
\end{subfigure}
\begin{subfigure}[t]{0.22\linewidth}
    \centering
\includegraphics[width=\linewidth]{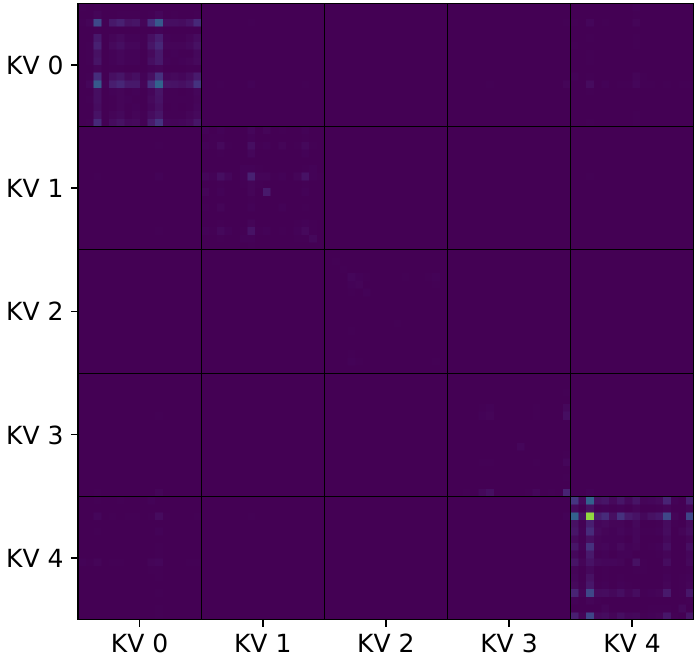}
    \caption{12000 Steps}
    \label{fig:ex2-hyperball-to-weight-decay}
\end{subfigure}
\begin{subfigure}[t]{0.22\linewidth}
    \centering
\includegraphics[width=\linewidth]{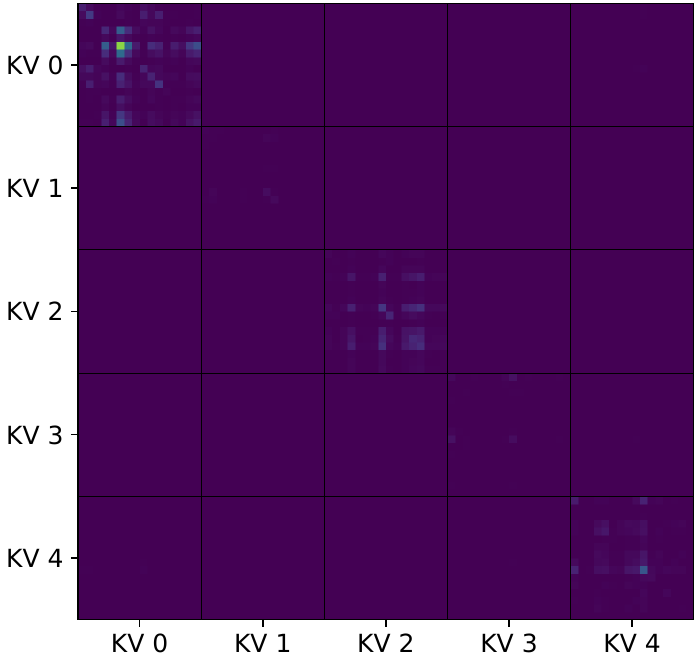}
    \caption{18000 Steps}
    \label{fig:ex2-hyperball-to-weight-decay}
\end{subfigure}
\begin{subfigure}[t]{0.265\linewidth}
    \centering
\includegraphics[width=\linewidth]{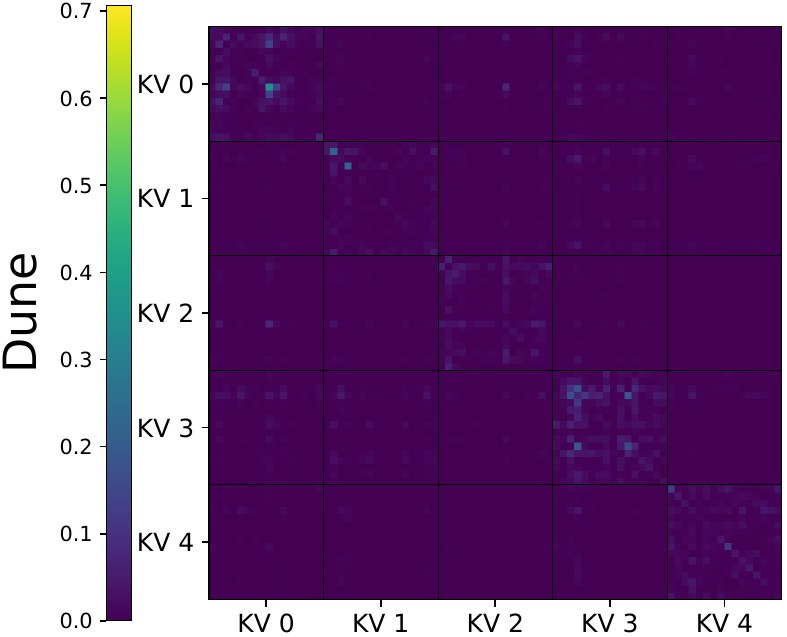}
    \caption{2000 Steps}
    \label{fig:ex2-hyperball-to-weight-decay}
\end{subfigure}
\begin{subfigure}[t]{0.22\linewidth}
    \centering
\includegraphics[width=\linewidth]{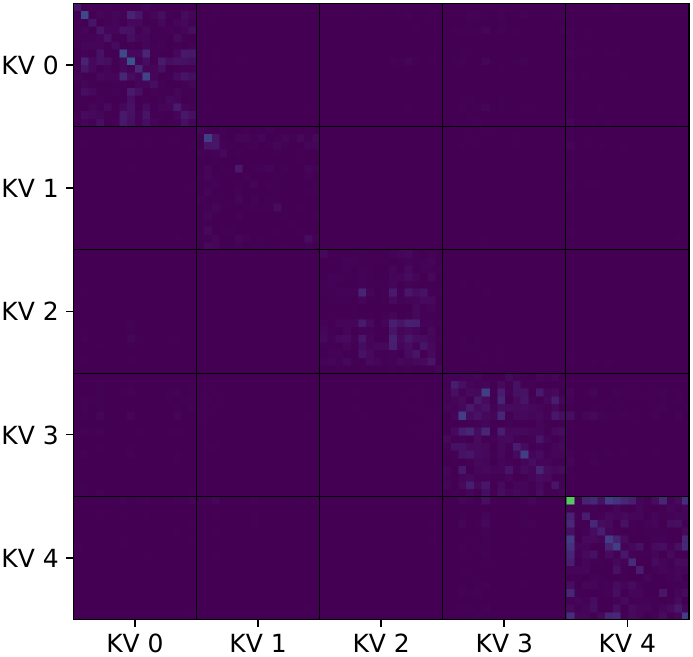}
    \caption{8000 Steps}
    \label{fig:ex2-hyperball-to-weight-decay}
\end{subfigure}
\begin{subfigure}[t]{0.22\linewidth}
    \centering
\includegraphics[width=\linewidth]{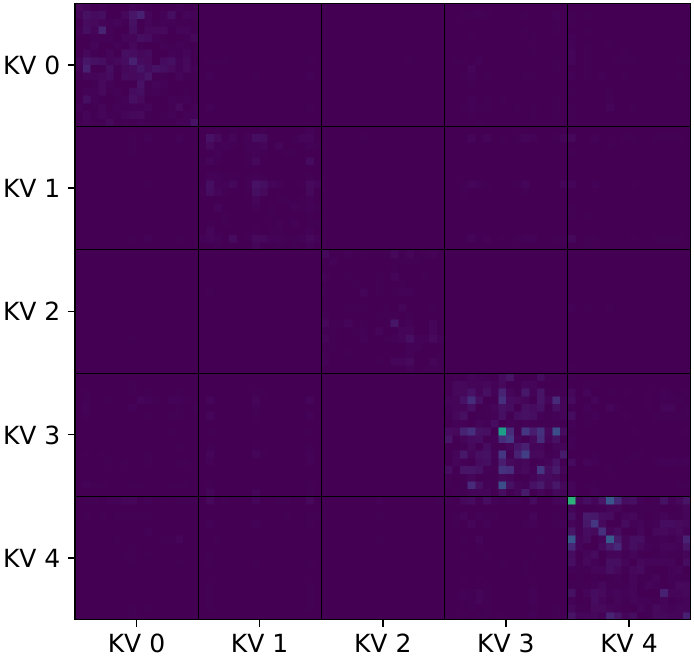}
    \caption{12000 Steps}
    \label{fig:ex2-hyperball-to-weight-decay}
\end{subfigure}
\begin{subfigure}[t]{0.22\linewidth}
    \centering
\includegraphics[width=\linewidth]{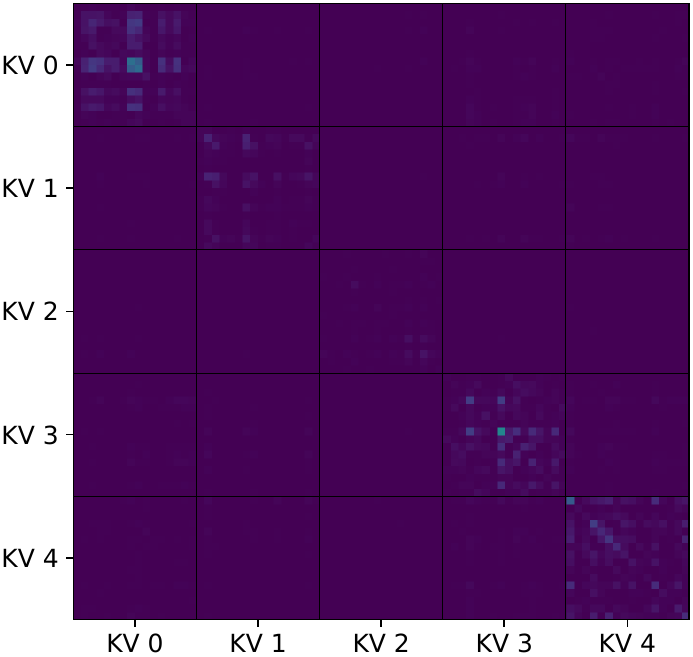}
    \caption{18000 Steps}
    \label{fig:ex2-hyperball-to-weight-decay}
\end{subfigure}
\caption{Evolution of block-diagonal curvature structure in Adam, Dust, and Dune during training.}
\label{fig:learning-444}
\end{figure}

\begin{figure}[t]
    \centering
    \includegraphics[width=\linewidth]{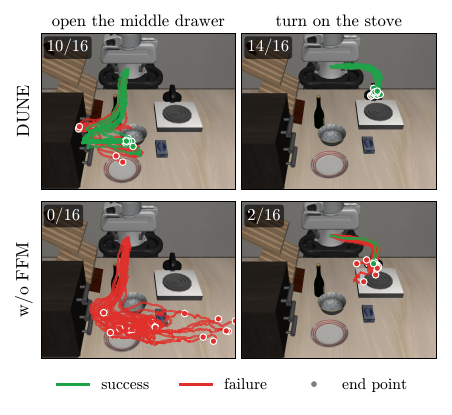}
    \caption{Closed-loop end-effector trajectories after independently fine-tuning the pretrained SmolVLA model for 12{,}000 steps in each setting. Each task starts from a fixed initial state (state 1 for the drawer task and state 6 for the stove task). We evaluate each policy with 16 samples of flow-matching initial noise per task. Green and red trajectories indicate successful and failed trials, respectively. Dots mark trajectory endpoints. The top row shows the full \textbf{Dune} model. The bottom row shows the ablation without time-conditioned historical regression.}
    \label{fig:fan}
\end{figure}

\subsection{The Robustness of Dune}
\textbf{Effectiveness of Time-Conditioned Historical Regression.}
The dynamics of flow-matching action generation vary across timesteps. To evaluate the contribution of time-conditioned historical regression to \textbf{Dune}, we fine-tune the pretrained SmolVLA model with and without this historical regression module. Removing historical regression yields a baseline that relies solely on current zeroth-order perturbation observations. As shown in Figure~\ref{fig:fan}, historical regression exploits the structure in past residuals across flow-matching time buckets to produce more stable closed-loop trajectories, with fewer divergent motions and task failures. These results suggest that historical regression provides reliable update guidance without global backpropagation.

\textbf{Task recovery under external disturbances.}
We also evaluate how the policies respond to sudden changes in the environment. After the robot first completes the task, we manually reset the environment state, for example by turning the knob back to its original position. The policy then has 200 additional interaction steps to complete the task again. As shown in Figure~\ref{fig:recover}, the policy fine-tuned with historical regression adjusts smoothly to the abrupt state change and completes the task again. In contrast, the policy trained without historical regression stalls after the disturbance, consistent with a noisier update signal during adaptation.

\subsection{The GPU Memory of Dune}
ZO optimization drastically reduces memory consumption on resource-constrained devices by eliminating the storage of intermediate activation graphs required by BP. As shown in Table~\ref{tab:memory}, \textbf{Dune} incurs a negligible overhead of only 150\,MiB over Dust for maintaining the compact historical statistics . Its peak GPU memory footprint of 10{,}898\,MiB represents a 41.3\% reduction compared to BP-based AdamW (18{,}578\,MiB), enabling full online TTA fine-tuning on consumer-grade 11\,GB GPUs (e.g., NVIDIA GTX 1080 Ti or GTX TITAN X) without auxiliary memory-saving techniques such as gradient checkpointing or parameter offloading.

\begin{table}[t]
    \centering
    \caption{Comparison of peak GPU memory footprint and minimum hardware requirements during SmolVLA fine-tuning (perturbation population size $K=4$, batch size 64).}
    \label{tab:memory}
    \renewcommand{\arraystretch}{1.2}
    \small
    \setlength{\tabcolsep}{8pt}
    \begin{tabular}{@{}ccc@{}}
        \toprule
        Method & GPU Memory  & Minimum Compatible GPU \\
        \midrule
        AdamW & 18{,}578MiB & RTX 3090 (24\,GB) \\
        Dust & 10{,}748MiB & \textbf{GTX 1080 Ti / GTX TITAN X} \\
        \textbf{Dune} & 10{,}898MiB & \textbf{GTX 1080 Ti / GTX TITAN X} \\
        \bottomrule
    \end{tabular}
\end{table}

\subsection{Interpretability Analysis of Dune}
\textbf{Curvature structure.}
To examine how Dune's design affects optimization dynamics, we compare the curvature structures of Adam, Dust, and Dune in Figure~\ref{fig:learning-444}. Ideally, second-order or adaptively preconditioned updates exhibit a clear block-diagonal structure. Adam maintains a pronounced block-diagonal pattern across the matrix throughout training. In contrast, Dust only briefly displays a local block-diagonal structure early in training before its diagonal blocks rapidly fade, which aligns with the observed performance plateau. Dune preserves this block-diagonal structure much more clearly than Dust, yielding a curvature pattern that closely resembles Adam's. 

\begin{figure}[t]
    \centering
    \includegraphics[width=\linewidth]{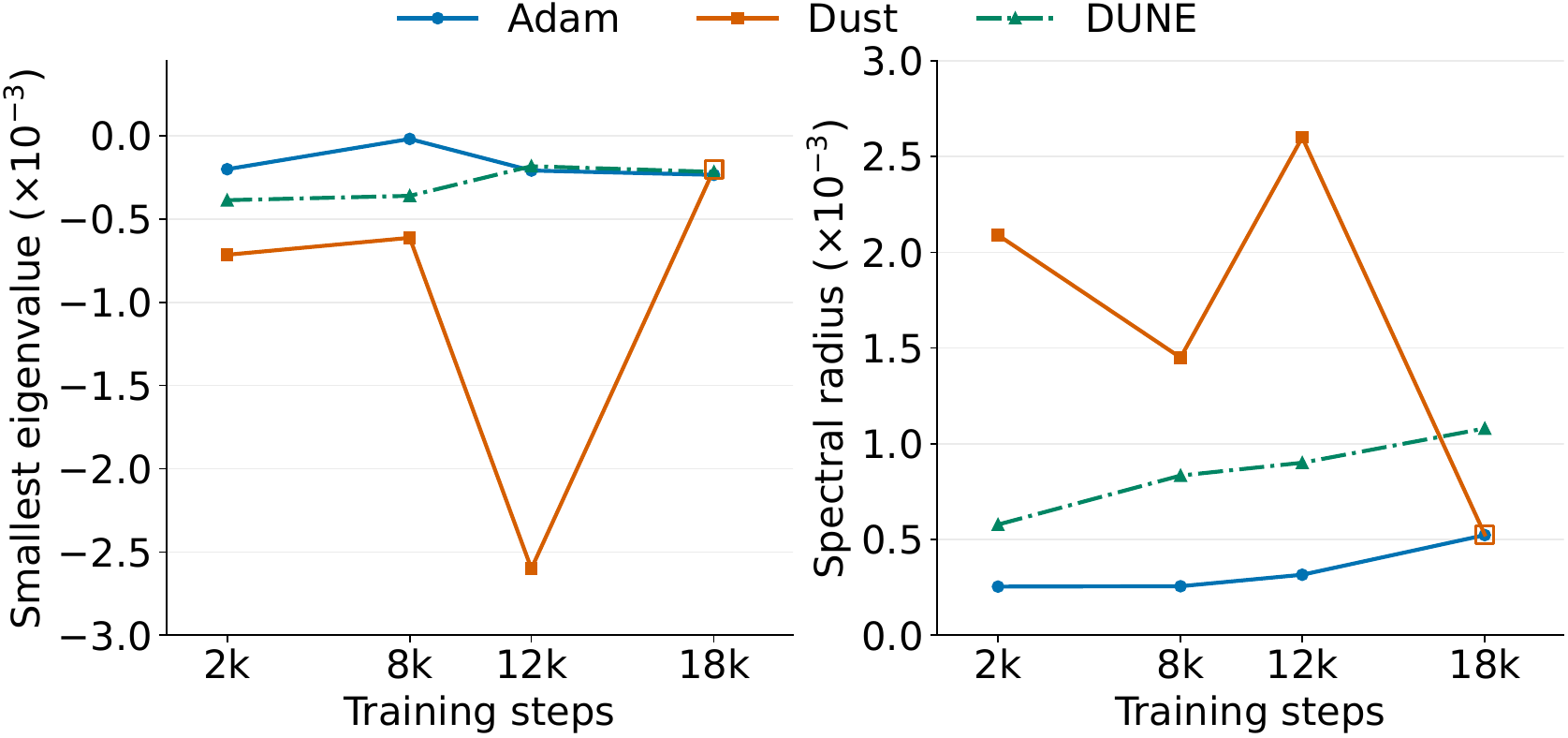}
    \caption{Evolution of Hessian spectral properties across training steps (2k--18k) for Adam, Dust, and Dune.}
    \label{fig:hessian_spectrum}
\end{figure}

Furthermore, Figure~\ref{fig:hessian_spectrum} tracks the evolution of Hessian spectral properties—specifically the smallest eigenvalue and spectral radius—across training steps (2k to 18k). Over long-horizon updates, \textbf{Dune} closely tracks Adam's smooth, nearly monotonic progression in both metrics, gradually mitigating negative curvature while keeping the spectral radius tightly bounded. Conversely, Dust suffers from severe, erratic fluctuations, exhibiting sharp spikes in spectral radius and deep drops into negative eigenvalues, which explains its optimization instability and performance degradation.

\section{Limitations}
The time--memory trade-off in ZO optimization illustrates the ``no free lunch'' principle. Substantial savings in GPU memory come at the cost of additional forward computation. As a result, ZO methods typically take longer to converge in wall-clock time than FO methods. Dune substantially reduces the GPU memory needed to store optimizer states, but running the same training workload with Dune takes several times as long as with Adam. This runtime gap is a key limitation of Dune and motivates further work on accelerating ZO gradient estimation in high-dimensional parameter spaces.
\section{Conclusion}
We propose Dune, a zeroth-order optimizer that fuses online historical error regression with current perturbation observations to mitigate performance plateaus during long-horizon VLA test-time adaptation. Conditioned on flow-matching time buckets, Dune estimates local error signals via decayed ridge regression without global backpropagation or additional forward passes, grounded in our theoretical analysis of variance-induced convergence bottlenecks. Experiments across LIBERO benchmarks and real-world robots show that Dune improves task success rates, closed-loop stability, and post-disturbance recovery under a low memory budget, demonstrating the strong potential of zeroth-order optimization for scalable, backpropagation-free adaptation in embodied AI.
\section*{LLM Usage Statement}
OpenAI services were used to assist with coding, and the accuracy of the resulting code was checked by the two co-first authors. Gemini was used to assist with language polishing, and the accuracy of the revised text was checked by the corresponding author.


{
    \small
    \bibliographystyle{ieeenat_fullname}
    \bibliography{main}
}


\end{document}